\documentclass[conference]{IEEEtran}
\IEEEoverridecommandlockouts
\usepackage{cite}
\usepackage{amsmath,amssymb,amsfonts}
\usepackage{algorithmic}
\usepackage{svg}
\usepackage{graphicx}
\usepackage{textcomp}
\usepackage{xcolor}
\usepackage{hyperref}

\def\BibTeX{{\rm B\kern-.05em{\sc i\kern-.025em b}\kern-.08em
    T\kern-.1667em\lower.7ex\hbox{E}\kern-.125emX}}
\begin{document}

\title{Care3D: An Active 3D Object Detection Dataset\\ of Real Robotic-Care Environments
}

\author{\IEEEauthorblockN{1\textsuperscript{st} Michael G. Adam}
\IEEEauthorblockN{2\textsuperscript{nd} Sebastian Eger}
\IEEEauthorblockN{3\textsuperscript{rd} Martin Piccolrovazzi}
\IEEEauthorblockN{10\textsuperscript{th} Eckehard Steinbach}
\IEEEauthorblockA{\textit{Chair of Media Technology}\\
\textit{Munich Institute of Robotics and Machine Intelligence (MIRMI)}\\
\textit{Department of Computer Engineering}\\
\textit{School of Computation, Information and Technology}\\
\textit{Technical University of Munich}\\
Munich, Germany\\
\{firstname\}.\{lastname\}@tum.de \vspace{1cm}}

\and
\IEEEauthorblockN{4\textsuperscript{th} Maged Iskandar}
\IEEEauthorblockN{5\textsuperscript{th} Joern Vogel}
\IEEEauthorblockN{6\textsuperscript{th} Alexander Dietrich}
\IEEEauthorblockN{11\textsuperscript{th} Alin Albu-Schaeffer}
\IEEEauthorblockA{\hspace{0.8cm}\textit{Institute of Robotics and Mechatronics}\hspace{0.8cm} \\
\textit{German Aerospace Center (DLR)}\\
Wessling, Germany \\
\{firstname\}.\{lastname\}@dlr.de}
\and
\hspace{1.5cm}
\and
\IEEEauthorblockN{7\textsuperscript{th} Seongjien Bien}
\IEEEauthorblockN{8\textsuperscript{th} Jon Skerlj}
\IEEEauthorblockN{9\textsuperscript{th} Abdeldjallil Naceri}
\IEEEauthorblockN{12\textsuperscript{th} Sami Haddadin}
\IEEEauthorblockA{\textit{Chair of Robotics and Systems Intelligence} \\
\textit{Munich Institute of Robotics and Machine Intelligence (MIRMI)}\\
\textit{Technical University of Munich}\\
Munich, Germany\\
\{s.bien, jon.skerlj, a.naceri, haddadin\}@tum.de}
\and
\hspace{0.5cm}
\and
\IEEEauthorblockN{13\textsuperscript{th} Wolfram Burgard}
\IEEEauthorblockA{\textit{Engineering Department} \\
\textit{University of Technology Nuremberg}\\
Nuremberg, Germany \\
0000-0002-5680-6500}
}

\maketitle

\begin{abstract}
As labor shortage increases in the health sector, the demand for assistive robotics grows. 
However, the needed test data to develop those robots is scarce, especially for the application of active 3D object detection, where no real data exists at all. This short paper counters this by introducing such an annotated dataset of real environments. The captured environments represent areas which are already in use in the field of robotic health care research. We further provide ground truth data within one room, for assessing SLAM algorithms running directly on a health care robot.
\end{abstract}


\begin{figure}%
    \centering
    \includegraphics[width=1\linewidth]{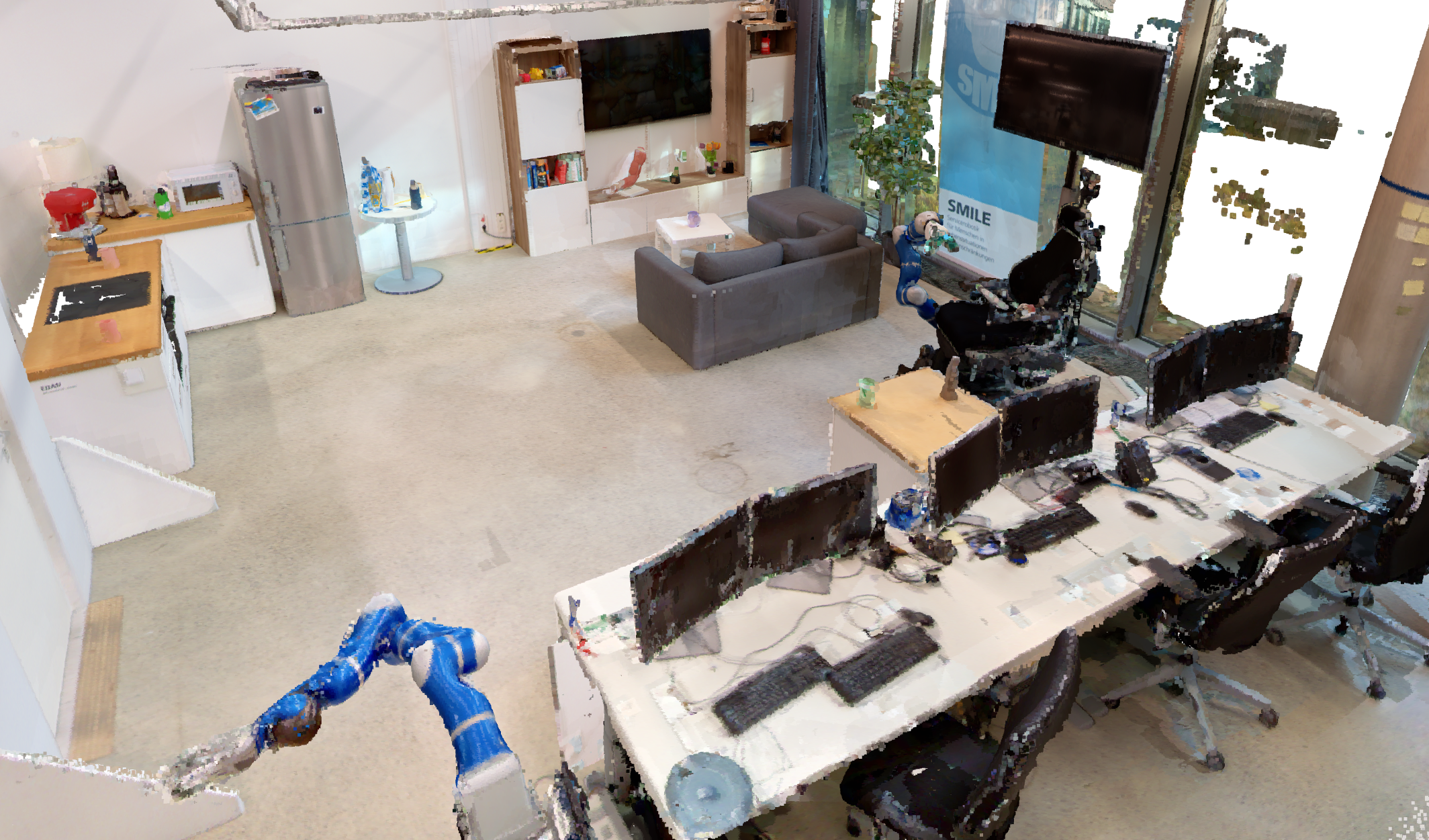}
    \caption{Rendered image of the captured scene at DLR.}
    \label{fig:render_dlr}
\end{figure}

\section{Introduction}
\begin{figure}%
    \centering
    \includegraphics[width=1\linewidth]{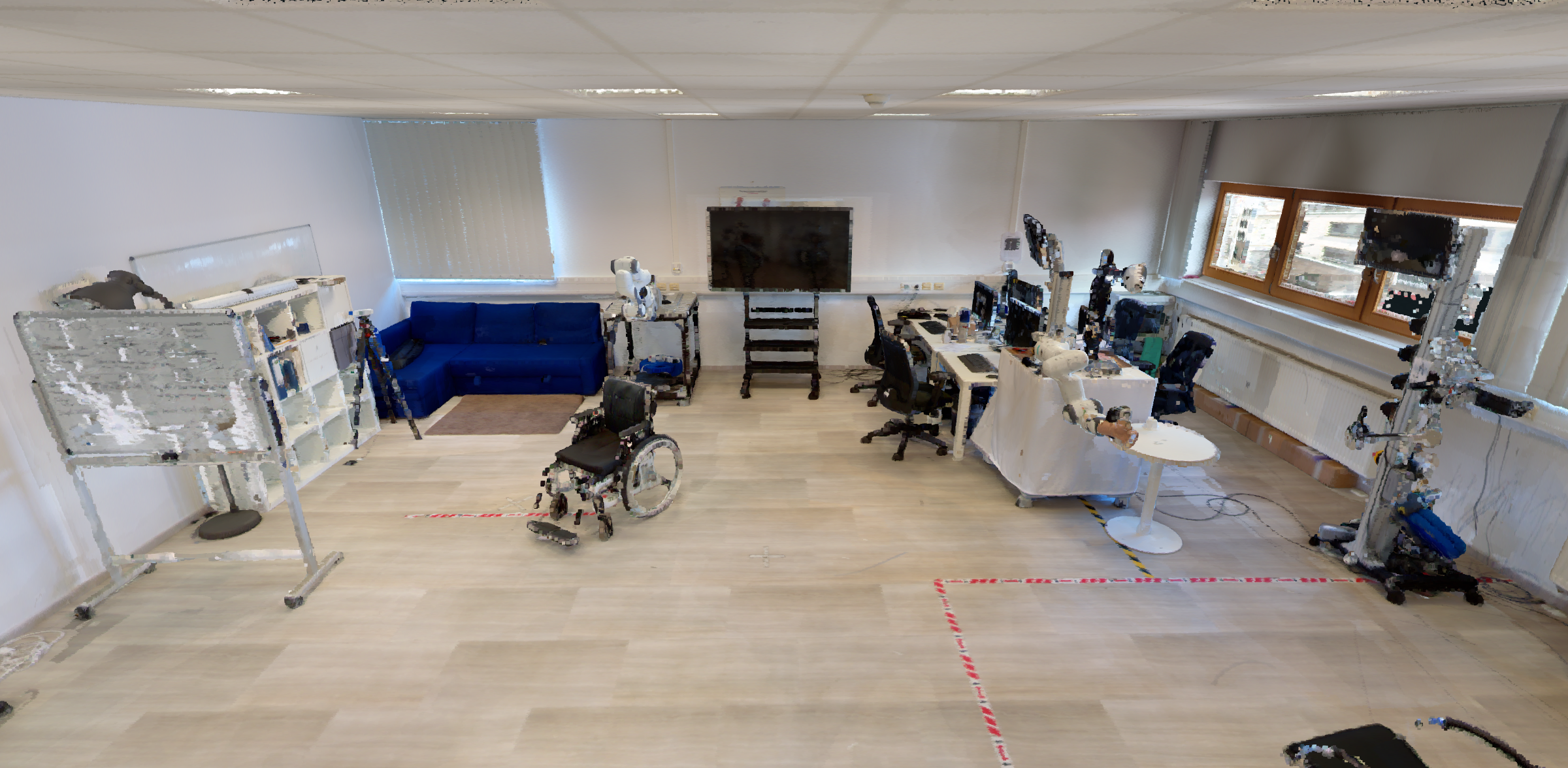}\\
    \vspace{0.1cm}
    \includegraphics[width=1\linewidth]{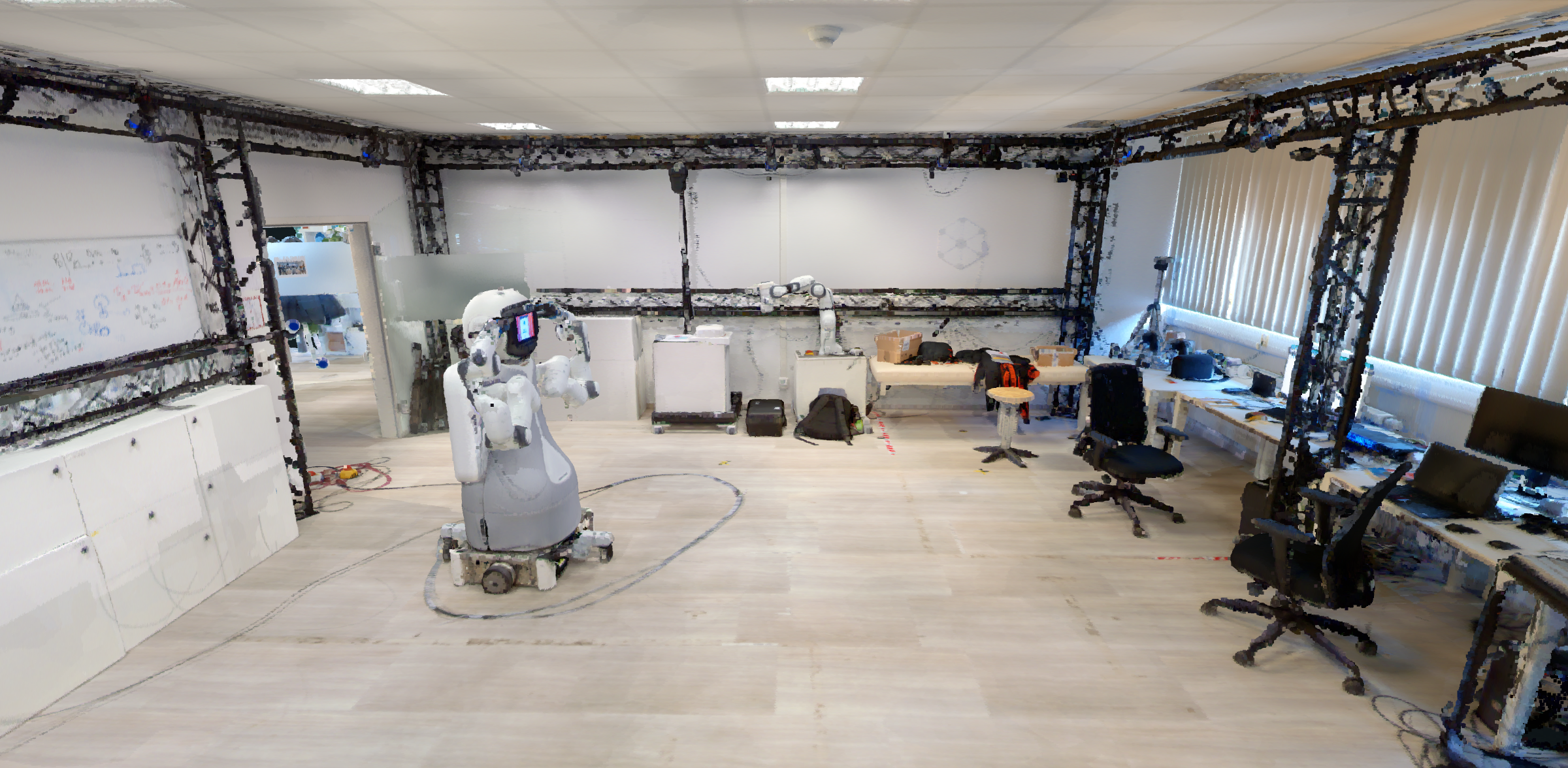}
    \caption{Rendered images of the scanned 3D scenes at GARMI.}
    \label{fig:render_garmi}
\end{figure}

Labor shortage in care facilities increases in Germany and other developed countries \cite{who:careshortage}. This has several reasons, such as bad working conditions and the aging of the society due to medical advances and due to demographic changes. In order to counter this issue, several solutions have been proposed by politics \cite{politic:healtcare1,politic:healtcare2}. One of them is the integration of robots into the care process \cite{robots:careshortage}.\\
However, to be able to support in care environments, the robots have to understand their surroundings. This can be achieved, among others, by advanced 2D and 3D object detection methods \cite{paper:objdect1,paper:objdect2,paper:objdect3}. Most state-of-the-art methods require training data, usually specialized on the application domain. This training data usually is either simulated or hard to get, expensive to annotate and only usable for static object detection.\\
In robotics physics-based simulations can help to adjust the control algorithms. Applying those algorithms on hardware leads to the so-called sim-to-real gap, yielding errors in real life, which would not occur in the simulator. This gap is harder to close the more complex/higher dimensional the simulated information is. An example of this is the creation of visual data. Rendering real-looking environments from virtual data is still a challenging problem, even though advances are made in this direction, such as real-time ray-tracing \cite{software:omniverse}.\\
Nevertheless, for high accuracy, fine tuning on real data is always needed. For the purpose of active object detection a whole real scene needs to be captured, which further allows for movement in the area. In this paper we publish a new dataset which is in the domain of health care. The set consists of two scenarios. One was captured at the German Aerospace Center (DLR), the other at the lighthouse initiative at Munich Institute of Robotics and Machine Intelligence (MIRMI), others will follow. Both scenes are shown in Fig. \ref{fig:render_dlr} and Fig. \ref{fig:render_garmi}. The data is available through a dedicated github project\footnote{\url{https://github.com/M-G-A/Care3D}}.\\

\section{Related Work}
3D object detection datasets, such as \cite{paper:scannet,paper:stanford,paper:matterport,paper:arkitscenes}, usually  only capture a small set (1-4) of photos within the scenes. Due to this sparsity, developing active visual algorithms has to deal additionally with the challenging task of rendering realistically looking photos from the meshes or point clouds. Those renders usually posses errors and introduce a sim-to-real gap.\\
Piccolrovazzi et al. \cite{paper:aTUM3D} introduce a solution in which real environments are captured. It allows for discrete movement by capturing $360^\circ$ panoramic images in a grid pattern across the room. Algorithms, such as active object detection, then can choose the next view from a finite set of possible positions.  We follow the same capturing process and our data shares the same structure. In such a way the algorithms which were developed for \cite{paper:aTUM3D} are compatible with this dataset. \\

\section{Dataset}
\subsection{Scanning Process}
The data was captured using a NavVis VLX~\cite{product:navvis} directly delivering $360^\circ$ with a resolution of $8192 \times 4096$. The resulting dense point cloud has an accuracy of at least $5\,mm$. During capture the position of the panoramas is registered as well.\\
The 3D bounding boxes with full degree-of-freedom were labeled in an adopted version of potree\cite{paper:potree}.

\subsection{DLR}
\begin{figure}%
    \centering
    \includegraphics[width=0.8\linewidth]{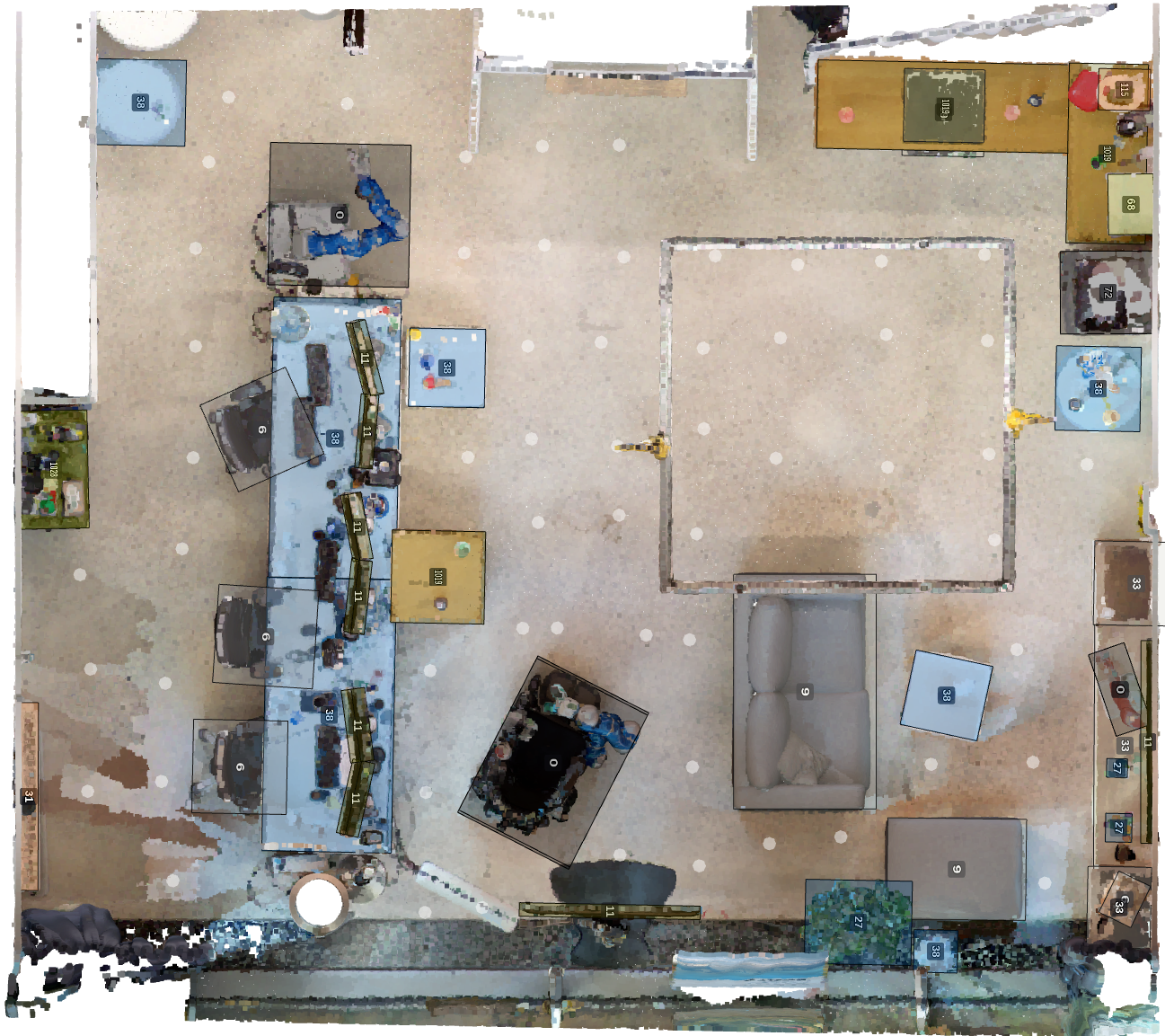}
    \caption{Overview of the scene captured at DLR. The boxes indicate the labeled objects. The white spheres show the position of captured panoramic images.}
    \label{fig:map_dlr}
\end{figure}
\subsubsection{Description}
The recording shows the SMiLE scenario, which serves as a test site for assistive robotics in care at DLR. The room contains relevant elements of a small apartment, including a standard door, a kitchen with cabinets, fridge, stove and oven, and a living room with TV, cupboards and couch. At DLR, this setting is used to investigate how assistive robotic systems can support people with disabilites in everyday activities \cite{vogel2020ecosystem}. 
The robots can be intuitively commanded by the users at various levels of autonomy, ranging from direct teleoperation via shared-control to supervised autonomy \cite{9341156}. Alternatively, an expert can teleoperate the robot from remote, in case of medical emergencies, or when the robot faces technical problems.
\begin{figure}%
    \centering
    \includegraphics[width=1\linewidth]{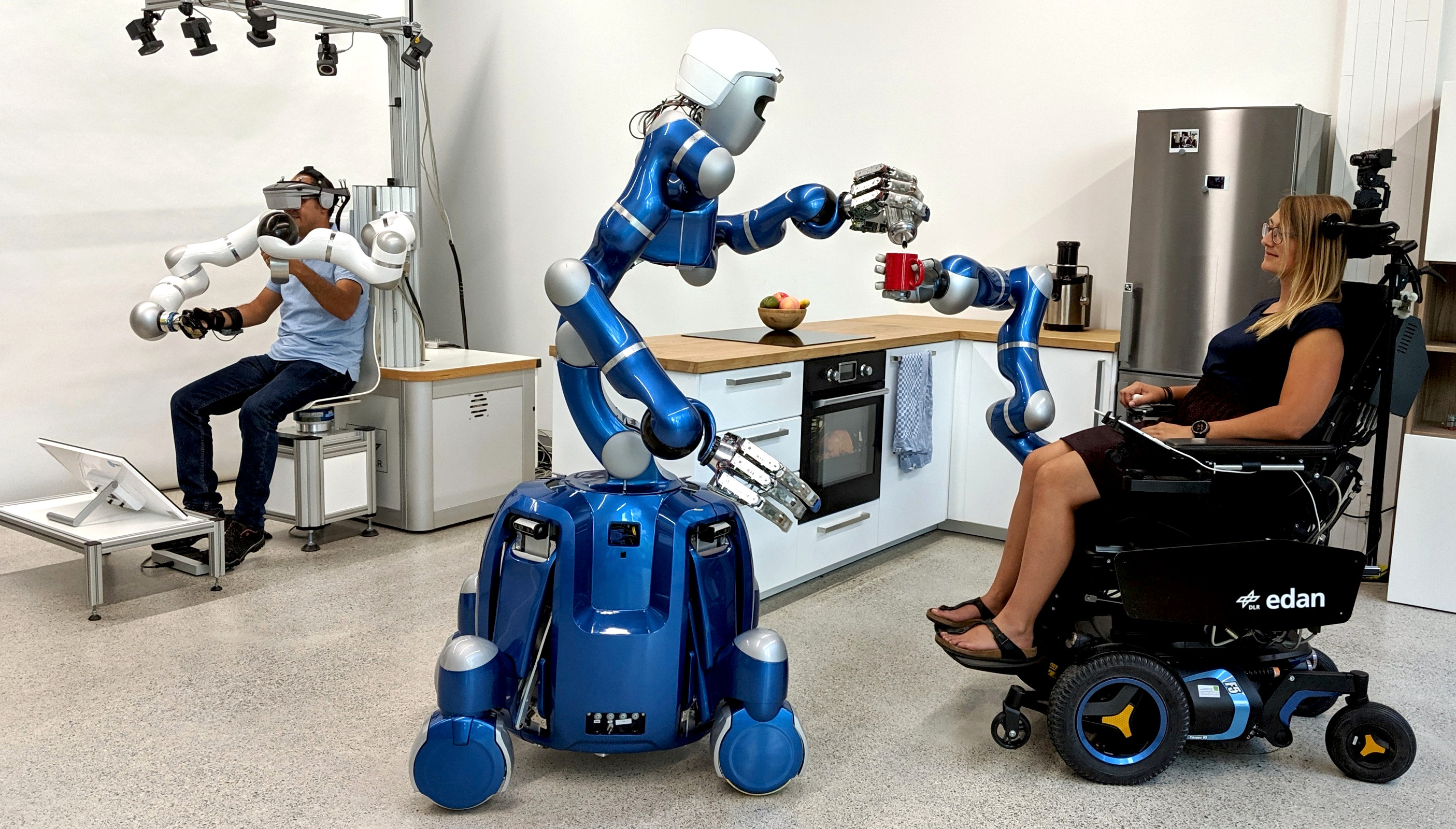}
    \caption{In the SMiLE Ecosystem we envision heterogeneous robotic assistants to support people with disablity. To increase dependability of the systems, remote haptic teleoperation can be used in case of technical issues or medical emergencies.}
    \label{fig:stat}
\end{figure}

\subsubsection{Stats}
The captured area covers $57.24\,m^2$. The robot has a navigatable space of $49.44\,m^2$ (calculated according to \cite{paper:CASE}). $58$ panoramic images have been captured. Their positions can be seen in Fig.~\ref{fig:map_dlr}. The same figure further depicts the location of the $40$ objects which have been annotated.

\subsection{GARMI}
\begin{figure}%
    \centering
    \includegraphics[width=1\linewidth]{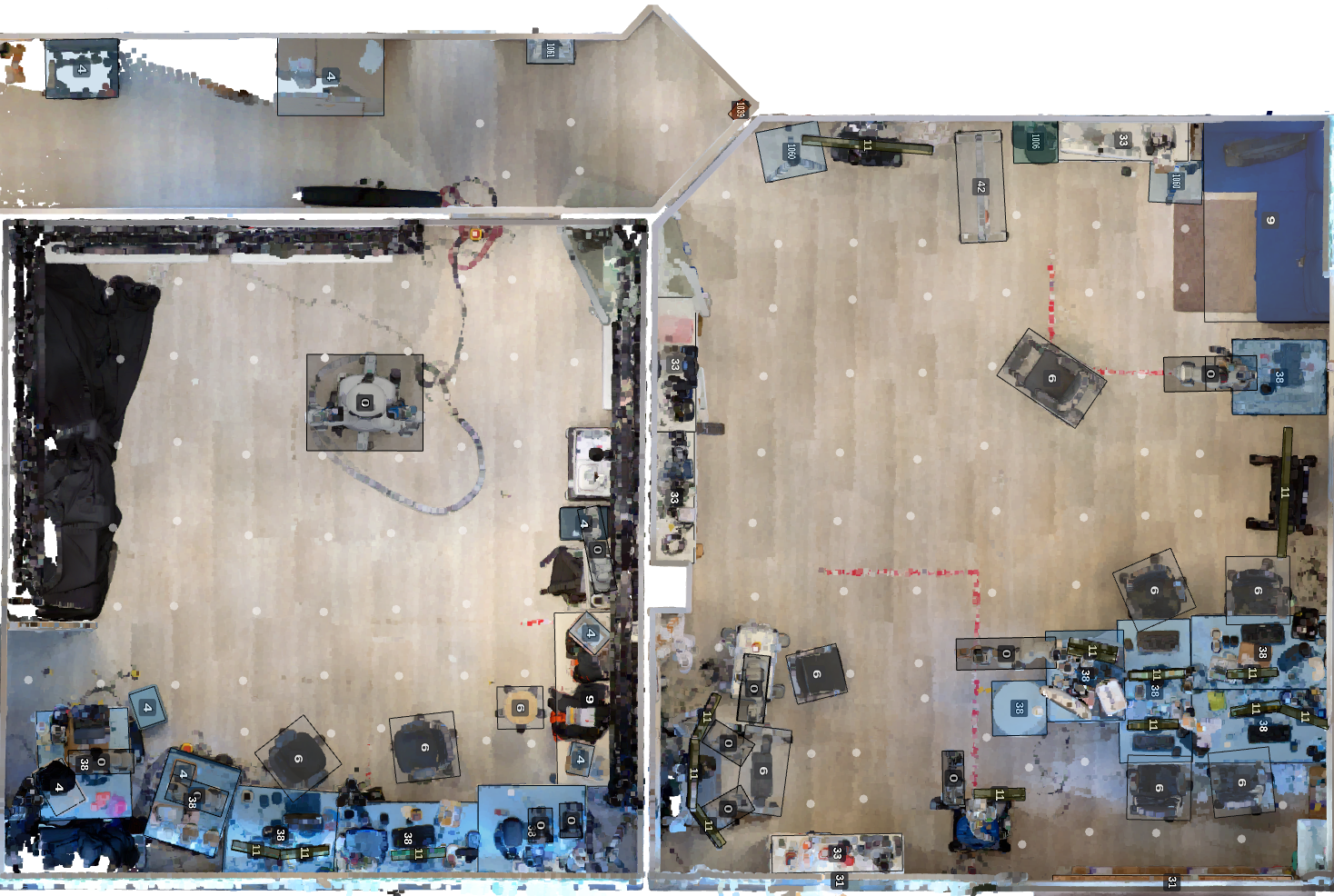}
    \caption{Overview of the scene captured with GARMI. The boxes indicate the labeled objects.  The white spheres show the position of captured panoramic images.}
    \label{fig:map_garmi}
\end{figure}

\subsubsection{Description}
The second scenario was recorded at the geriatronics lighthouse project of TUM. The project goal is to develop a humanoid robot, called GARMI. GARMI will help in the rehabilitation phase by supporting patients physically. Further, it gives a tele-operation platform for doctor visits. The scene consists of two rooms. One room mimics a doctor office like surrounding, equipped with tele-operatable robotic arms and the second is showing GARMI itself in a controlled development environment.\\
GARMI is a bimanual service robotics platform, built on a holonomic base with two actuated wheels.
The robot's locomotion controller has been developed to counteract the inertial swinging of the robot body caused by acceleration and deceleration.
The robot is equipped with two Intel Realsense D435(i) cameras, one IMU model on the forehead and one non-IMU model in the torso. Both cameras were calibrated using the vendor-provided calibration software package before the recording. 

\begin{figure}%
    \centering
    \includegraphics[width=1\linewidth]{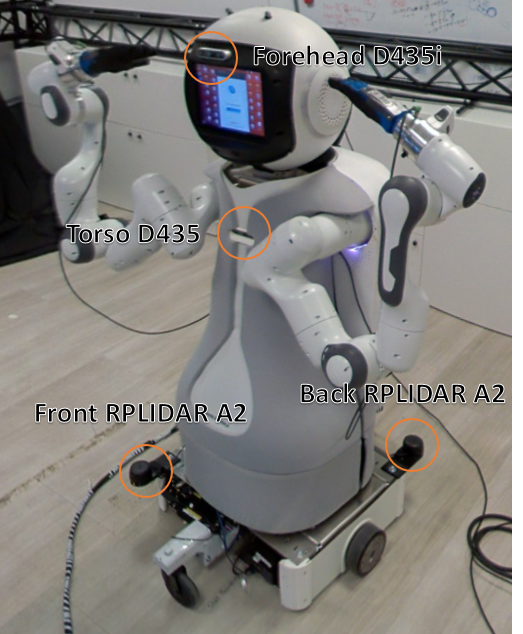}\\
    \caption{Location of sensors on GARMI.}
    \label{fig:garmi_setup}
\end{figure}

\subsubsection{Stats}
Two rooms connected through an attached floor have been captured. The area is $138.36\,m^2$ big, of which $129.43\,m^2$ is accessible by the robot. The space has $107$ discrete image positions, which also allow for movement between the rooms. $70$ objects have been annotated. The map is shown in Fig.~\ref{fig:map_garmi}.

\subsubsection{SLAM Test-Set}
We additionally captured sensor data of the GARMI robot moving in 2 different patterns across the second room. This data can be used to assess a visual SLAM algorithm, as ground truth position and pose was captured by a VICON-system, which is installed across the ceiling. Further, the point cloud generated by the VLX can be used as a ground truth for assessing the reconstruction capabilities of the algorithm, for instance by comparing the resulting point cloud with methods such as ICP~\cite{icp,icp2}.\\

\begin{figure}%
    \centering
    \includegraphics[width=1\linewidth]{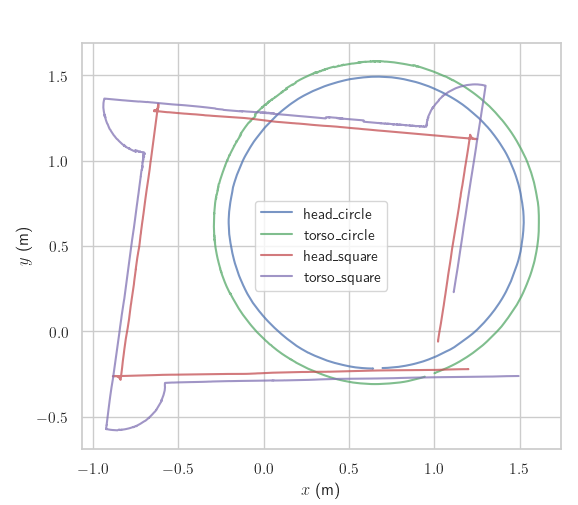}
    \caption{The two patterns used while recording the SLAM Test-Set.}
    \label{fig:traj}
\end{figure}

In some of the recordings, the head of the robot is actuated to allow for pitch and yaw rotations. The robot's two arms were positioned to remove any occlusion from both of the cameras.
Additionally, the robot is equipped with two RPLIDAR-A2 scanners in the front and back side of the base.
The overall sensor setup, as well as the robot's arm joint positions throughout the recording can be seen in Fig. \ref{fig:garmi_setup}.
The recorded trajectory patterns can be seen in Fig.~\ref{fig:traj}.

In this subset, the following data types are available:
\begin{itemize}
    \item The RGBD images and camera information of the two cameras
    \item IMU data from the forehead-mounted D435i camera
    \item Raw LIDAR data from the two RPLIDAR-A2s
    \item VICON tracking data for ground truth pose estimation of the head and torso
\end{itemize}


\subsection{Label}

\begin{figure}%
    \centering
    \includegraphics[trim=0 5.1cm 0 0, clip, width=1\linewidth]{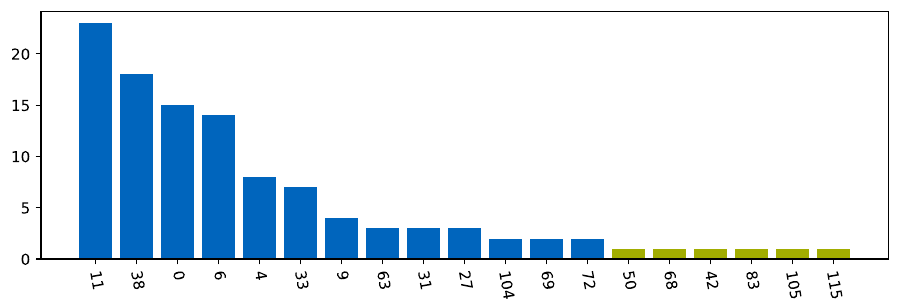}
    \caption{Histogram of the class occurrences. The green classes occur only once and can be merged to one 'other' class.}
    \label{fig:stat}
\end{figure}
As in \cite{paper:aTUM3D} the class distribution of the labels is highly nonlinear as visualized in Fig.~\ref{fig:stat}. The bounding boxes posses full degree of freedom, such that rotation around any axis is allowed. The average edge length of the objects is $0.71\,m$. The average volume is $0.45\,m^3$.

\subsection{Conclusion}\label{AA}
In this paper we shortly introduce a new real world dataset for active 3D object detection in the field of health care. It follows the same structure as \cite{paper:aTUM3D} and further allows the assessment of a SLAM system running on an assistant robot. More scenarios will be added in the future.

\bibliographystyle{IEEEtran}

\bibliography{literature}

\vspace{12pt}
\color{red}

\end{document}